\documentclass{edm_article}

\usepackage{graphicx}
\usepackage{subfigure}
\usepackage{caption}

\usepackage[linesnumbered,ruled,vlined]{algorithm2e}
\usepackage{booktabs}
\usepackage{subcaption}
\usepackage{float}

\begin{document}

\title{Toward Sufficient Statistical Power in Algorithmic Bias Assessment: A Test for ABROCA}

\numberofauthors{1}
\author{
Conrad Borchers\\
       \affaddr{Carnegie Mellon University}\\
       \email{cborcher@cs.cmu.edu}
}

\maketitle

\begin{abstract} 
Algorithmic bias is a pressing concern in educational data mining (EDM), as it risks amplifying inequities in learning outcomes. The Area Between ROC Curves (ABROCA) metric is frequently used to measure discrepancies in model performance across demographic groups to quantify overall model fairness. However, its skewed distribution--especially when class or group imbalances exist--makes significance testing challenging. This study investigates ABROCA's distributional properties and contributes robust methods for its significance testing. Specifically, we address (1) whether ABROCA follows any known distribution, (2) how to reliably test for algorithmic bias using ABROCA, and (3) the statistical power achievable with ABROCA-based bias assessments under typical EDM sample specifications. Simulation results confirm that ABROCA does not match standard distributions, including those suited to accommodate skewness. We propose nonparametric randomization tests for ABROCA and demonstrate that reliably detecting bias with ABROCA requires large sample sizes or substantial effect sizes, particularly in imbalanced settings. Findings suggest that ABROCA-based bias evaluations based on sample sizes common in EDM tend to be underpowered, undermining the reliability of conclusions about model fairness. By offering open-source code to simulate power and statistically test ABROCA, this paper aims to foster more reliable statistical testing in EDM research. It supports broader efforts toward replicability and equity in educational modeling.
\end{abstract}


\section{Introduction}

Algorithmic bias is a critical concern in educational data mining (EDM), as it can undermine the effectiveness of key EDM applications. Past research has studied algorithmic bias in knowledge tracing \cite{zambrano2024investigating}, grade prediction \cite{jiang2021towards,deho2024past,vsvabensky2024evaluating}, and at-risk prediction \cite{cock2023protected,karimi2021predicting}. Despite differences in definitions \cite{baker2022algorithmic}, algorithmic bias generally refers to systematic unfairness or unequal treatment caused by algorithms, often disadvantaging or favoring certain individuals or groups of individuals in favor of others \cite{friedman1996bias}. In this context, efforts to measure algorithmic fairness in EDM models have often focused on metrics that quantify performance differences between student groups based on demographic attributes (e.g., by race or gender). A model is typically considered fair if its predictive performance is equivalent or comparable across groups.

However, like all statistical metrics, measures of algorithmic bias are subject to random fluctuation through sampling. A critical yet underexplored dimension in algorithmic bias research is the role of statistical power in bias detection. Statistical power represents the probability of correctly rejecting the null hypothesis—concluding a model is biased when it truly is. Inferential conclusions about model fairness are vulnerable to both Type I (false positive) and Type II (false negative) errors. For example, insufficient statistical power might lead to the erroneous conclusion that a model is fair, masking genuine disparities in performance among demographic groups (false negative). Conversely, a model may be incorrectly deemed biased when it is not (false positive). While considerations of statistical power are widely recognized in other EDM applications \cite{haim2023open}, prior research reporting that models are fair have typically not considered statistical power \cite{jiang2021towards,zambrano2024investigating,borchers2025abroca}. Furthermore, systematic investigations into how sample sizes, effect sizes, and design factors influence the reliability of bias detection remain scarce. One exception is Choi et al. \cite{choi2025bias}, who applied bootstrapping to a dropout prediction data set to quantify uncertainty in bias metrics through confidence intervals. This gap highlights an urgent need to incorporate robust power analysis and reporting standards to ensure fairness claims—or their absence—are replicable, meaning they are generalizable to other samples \cite{haim2023open}.

Among the various metrics for assessing algorithmic bias, ABROCA is an illustrative example that underscores the importance of statistical power considerations. ABROCA measures discrepancies in classification performance between demographic groups by comparing the area under their respective receiver operating characteristic (ROC) curves. Larger ABROCA values indicate greater performance disparities, signaling substantial algorithmic bias. However, recent studies have highlighted the skewed distribution of ABROCA, particularly in datasets with class or group imbalances \cite{borchers2025abroca}. In other words, large ABROCA values may arise purely by chance, even when no population-level bias or differences in ROC curves exist. This can lead to false-positive conclusions about a model’s bias based on observed ABROCA values driven by random sampling variability.

The susceptibility of ABROCA to producing misleading signals of bias necessitates more rigorous statistical testing. Researchers seeking to assess the significance and reliability of observed ABROCA differences face several challenges. How can they reliably evaluate if observed discrepancies are due to genuine bias or chance? How can they mitigate the risk of false positives, especially given the wide variation in sample sizes and group distributions typical of EDM studies \cite{vsvabensky2024evaluating,zambrano2024investigating}? At present, no established statistical test determines whether an observed ABROCA value reflects significant bias, partly because the distribution of ABROCA remains poorly understood \cite{borchers2025abroca}. 

This paper addresses these challenges by developing a method to statistically test ABROCA in its most commonly used form, that is, as a metric for overall algorithmic fairness. Specifically, we contribute methods and empirical insights into the role of statistical power in algorithmic bias assessment. These statistical testing and power simulation methods may extend to other metrics commonly used and developed in EDM \cite{verger2023your,xu2024contexts}. We focus on the following research questions:

\textbf{RQ1:} What statistical test is suitable for assessing significant algorithmic bias through ABROCA's distribution?\\
\textbf{RQ2:} How much statistical power does this test achieve under typical EDM study sample sizes?\\
\textbf{RQ3:} How does the statistical power of ABROCA-based bias assessment depend on group and outcome class imbalances?

By addressing these questions, we provide practical guidance for researchers on selecting statistical methods to test bias using ABROCA while minimizing false positives and ensuring reliable claims of fairness. To promote reproducibility, we contribute open-source code for simulating statistical power and testing ABROCA under varying conditions.\footnote{https://github.com/conradborchers/abroca-test-power} This work enhances the rigor of algorithmic fairness research in EDM, contributing to emerging fairness evaluation methods.

\section{Related Work}

\subsection{What is Algorithmic Bias?}

Algorithmic bias is a critical issue in EDM, where fairness in model performance across diverse student populations is essential to ensure equitable educational outcomes. Broadly defined, algorithmic bias refers to systematic and unfair discrimination against specific individuals or groups, often favoring others \cite{friedman1996bias}. In the context of education, such bias can exacerbate existing inequities, undermining the potential of data-driven interventions to support all learners effectively.

There are three primary paradigms for understanding algorithmic bias in education: similarity-based fairness, causal fairness, and statistical fairness \cite{verma2018fairness,kizilcec2022algorithmic,barocas2020hidden}. 

\textbf{Similarity-based fairness} focuses on ensuring that students with similar profiles—those sharing comparable feature distributions—are treated equivalently by the model. For example, if two students possess equivalent prior knowledge and engagement levels, their predicted learning outcomes should be similar regardless of demographic differences.

\textbf{Causal fairness} evaluates fairness by modeling the causal relationships between sensitive attributes (e.g., race or gender) and predictions. Through counterfactual reasoning, causal fairness examines whether changing a student's sensitive attribute would lead to a different prediction. If a prediction changes solely due to altering a sensitive attribute, the model would be considered unfair under this paradigm \cite{verma2018fairness}.

\textbf{Statistical fairness} assesses fairness by examining aggregate performance metrics across demographic groups. Metrics such as overall accuracy are compared to determine whether a model's performance is equitable across groups. Statistical fairness--overall fairness assessed through comparison of model accuracies on subgroups--presents the most common paradigm of algorithmic fairness in EDM \cite{verger2023your,zambrano2024investigating,vsvabensky2024evaluating}, including the most common use of ABROCA \cite{gardner2019evaluating,borchers2025abroca}. Therefore, the present study is also situated in the statistical fairness paradigm.

\subsection{Algorithmic Bias Assessment in EDM}

To measure bias, in recent years, the field of EDM has seen a surge in the development of novel metrics and methodologies, such as equalized odds \cite{xu2024contexts}, the Model Absolute Density Distance (MADD) \cite{verger2023your}, pseudo $R^2$ to predict the correctness of individual predictions based on demographic group attributes \cite{xu2024contexts}, and ABROCA \cite{gardner2019evaluating}. These measures are typically used to determine \textit{if} a model is biased at all, or if a model is biased than another model \cite{xu2024contexts}, for example, in the context of reducing a model's bias through different training regimes \cite{cock2023protected}. Each of these metrics assumes that bias can vary--based on these metrics--and that researchers can conclude that a model is biased or not (or more biased than another model) by inspecting and comparing these metrics.

\subsection{The ABROCA Metric}

A widely used measure for evaluating binary classifiers is the area under the ROC curve (AUC) metric \cite{bowers2019receiver}. The AUC metric offers an intuitive interpretation (the probability that the classifier correctly distinguishes a positive instance from a negative one) and reliable performance assessment for imbalanced class distributions \cite{jeni2013facing}. Its robustness stems from evaluating the true positive rate (TPR) and false positive rate (FPR) across all classification thresholds from 0 to 1. The ROC curve plots TPR against FPR at varying thresholds, and the AUC quantifies the area beneath this curve. A higher AUC indicates better performance, with 0.5 representing chance-level prediction and 1 denoting perfect classification.

ABROCA is a widely adopted fairness metric that measures disparities in classifier performance across demographic subgroups. Introduced by Gardner et al. \cite{gardner2019evaluating}, it quantifies biases along the decision-threshold spectrum by computing the area between subgroup-specific ROC curves. Unlike traditional AUC-based comparisons, which assess overall performance, ABROCA identifies localized disparities—revealing where a model may favor one group over another at specific thresholds. This granular approach makes ABROCA particularly valuable for detecting nuanced fairness issues that aggregate metrics might overlook.

The utility of ABROCA has been demonstrated in various studies. Xu et al. \cite{xu2024contexts} applied ABROCA to quantify fairness in the context of predictive model transfer for higher education academic performance prediction. Their study employed the Wilcoxon signed-rank test to compare ABROCA values between two models applied to the same dataset, highlighting that ABROCA is often used in a significance testing framework (or at least with the intention of statistical comparison). Similarly, Sha et al. \cite{sha2021assessing} utilized ABROCA to investigate gender differences in classifying educational forum posts, revealing disparities in how models perform across male and female student contributions. Deho et al. \cite{deho2024past} used ABROCA to examine the impact of dataset drift on fairness in course grade prediction models. Their findings underscored the susceptibility of fairness metrics, including ABROCA, to changes in data distributions over time.

These studies collectively demonstrate ABROCA's versatility in measuring fairness across various contexts in EDM. They also highlight the importance of understanding the metric’s statistical properties, as ABROCA is often used to assess a classifier's overall algorithmic fairness, including through methods of statistical comparison and significance testing.

\section{Method}

\subsection{RQ1: ABROCA's Distribution}
To investigate whether ABROCA follows a known distribution, we generated a simulated dataset of ABROCA values across various conditions, including variations in sample size, class imbalance ratios, and outcome imbalance ratios, following Borchers and Baker \cite{borchers2025abroca}. This investigation was conducted under the null hypothesis that both groups share the same AUC values and that the model is equally accurate for both groups. Simulations were performed under two experimental conditions: (1) balanced data, where both groups and outcome classes were of equal size, and (2) imbalanced data, with a 90\% imbalance ratio for both groups and outcomes. For this distributional analysis, 5,000 observations were simulated from each distribution.

We evaluated the distributional fit of ABROCA values by testing them against several theoretical distributions using the \texttt{fitdistrplus} package in R \cite{fitdistRplus}. In addition to comparing ABROCA to standard test statistic distributions such as the normal, $t$-, and $F$-distributions, we also fit the simulated ABROCA values to Weibull distributions, which are commonly applied to positively skewed data \cite{pinho2012gamma}.

Goodness-of-fit was assessed through visual diagnostics--quantile-quantile (Q-Q) plots--to compare the observed ABROCA values with the theoretical quantiles of each distribution. We also conducted the Kolmogorov-Smirnov (K-S) test to compare the empirical cumulative distribution function of the ABROCA values with the theoretical distributions. These analyses aimed to determine whether ABROCA conforms to a known parametric distribution, which would facilitate the application of parametric statistical testing methods.

\subsection{RQ2: Simulating Statistical Power}

A central question of this study is whether the ABROCA test has sufficient statistical power when applied to sample sizes typical in EDM. Once statistical tests for ABROCA are established, their power can be simulated under varying sample size specifications.

Statistical power quantifies the likelihood of detecting a true effect when it exists, such as identifying bias in a model or comparing biases between models. In frequentist hypothesis testing, there are two hypotheses: the null hypothesis ($H_0$), which assumes no bias or no difference in model accuracy, and the alternative hypothesis ($H_1$), which assumes the presence of bias or differences in model accuracy. Errors can occur in this framework: a false positive error (Type I) occurs when $H_0$ is incorrectly rejected, and a false negative error (Type II) occurs when $H_1$ is incorrectly rejected. The probability of committing a Type II error is denoted as $\beta$, and statistical power is defined as $1 - \beta$. Thus, power represents the probability of correctly detecting bias when it is present in the population (Table \ref{tab:confusion_matrix}).

\begin{table}[h!]
\centering
\caption{Confusion matrix for statistical testing in algorithmic bias assessment.}
\label{tab:confusion_matrix}
\begin{tabular}{@{}lcc@{}}
\toprule
               & \textbf{Conclude No Bias} & \textbf{Conclude Bias} \\ \midrule
\textbf{Not Biased} & True Negative (1-$\alpha$)        & False Positive ($\alpha$)         \\
\textbf{Biased}     & False Negative ($\beta$)         & True Positive (1-$\beta$)         \\ \bottomrule
\end{tabular}
\end{table}

Statistical power increases with sample size and more precise estimates \cite{wickelmaier2021simulating}. Statistical tests and simulation procedures must be developed to evaluate power in the specific context of algorithmic bias analysis with ABROCA. This study introduces a statistical test for ABROCA, enabling power estimation for common sample sizes in the field (see Section \ref{sec:method:test} and Algorithm \ref{algo:random-test}).

\begin{algorithm}[htp]
\caption{Simulating ABROCA with Input: $AUC_1$, $AUC_2$, $N_{\text{total\_sample}}$, $n_{\text{iter\_test}}$, $n_{\text{iter\_power}}$, $\text{ratio\_pos\_case}$, $\text{test\_set\_size}$; Output: ABROCA Metric, $p$, and Power}
\begin{enumerate}
    \item Simulate data from two distributions corresponding to $AUC_1$ and $AUC_2$ as described in \cite{borchers2025abroca}.
    \item Perform randomization tests to compute \textit{p}-values:
    \begin{enumerate}
        \item Randomly permute group labels $n_{\text{iter\_test}}$ times.
        \item Calculate ABROCA for each permutation \cite{gardner2019evaluating}.
        \item Compute the \textit{p}-value as the proportion of permuted ABROCA values exceeding the observed ABROCA.
    \end{enumerate}
    \item Simulate statistical power:
    \begin{enumerate}
        \item Repeat the randomization test $n_{\text{iter\_power}}$ times.
        \item Compute the proportion of tests where \textit{p} $< 0.05$.
    \end{enumerate}
    \item Conduct power analysis by varying sample size:
    \begin{enumerate}
        \item Adjust $N_{\text{total\_sample}}$ across a range of values.
        \item Plot sample size against statistical power.
    \end{enumerate}
\end{enumerate}
\label{algo:random-test}
\end{algorithm}

The steps for computing statistical power follow the framework proposed by Wickelmaier \citeyear{wickelmaier2021simulating}. First, a minimally relevant effect size must be determined. For this study, we assume an effect size corresponding to different AUC differences ranging from 0.02 to 0.2 as a meaningful starting point while recognizing that different effect sizes may have varying practical implications. Using simulation methods outlined by Borchers and Baker \cite{borchers2025abroca}, we repeatedly generate sample datasets under specified conditions, apply the statistical tests, and analyze the resulting $p$-value distributions. The proportion of simulations where the $p$-value falls below a significance threshold (e.g., $\alpha = 0.05$) provides an estimate of the statistical power.

To simulate the statistical power for testing ABROCA, the following steps are taken:

1. Define the statistical model and effect of interest. For ABROCA, this includes selecting an appropriate statistical test, such as a nonparametric randomization test, and fixing parameters relevant to the data-generating process, such as group proportions and expected effect sizes (i.e., population-level AUC differences). Effect sizes include 0.02, 0.05, 0.1, 0.15, 0.20.\\
2. Generate observations under varying conditions, including sample size, class imbalance, and group proportions. These simulations use open-source code \cite{borchers2025abroca}.\\
3. Apply statistical tests to evaluate $H_0$, which posits no difference in ABROCA values across groups. The test should align with the intended analysis, in this case, testing if a model exhibits significant bias.\\
4. Repeat the process for multiple iterations. Each iteration produces a $p$-value, and the proportion of $p$-values below the threshold $\alpha = 0.05$ is the test's power.

The results provide insights into the ability of ABROCA-based statistical tests to detect true differences in bias. We evaluate test set sizes between 100 and 2,000 samples, which is typical in educational data mining research \cite{zambrano2024investigating,vsvabensky2024evaluating}.

\subsection{RQ3: Statistical Power and Imbalance}
\label{sec:method:test}

This methodology evaluates the statistical power of ABROCA-based bias assessments through simulations that account for variations in sample sizes, effect sizes, and group distributions. For details on the ABROCA simulation methodology, we refer to Borchers and Baker \cite{borchers2025abroca}.

\begin{figure}[htp]
    \centering
    \includegraphics[width=0.485\textwidth]{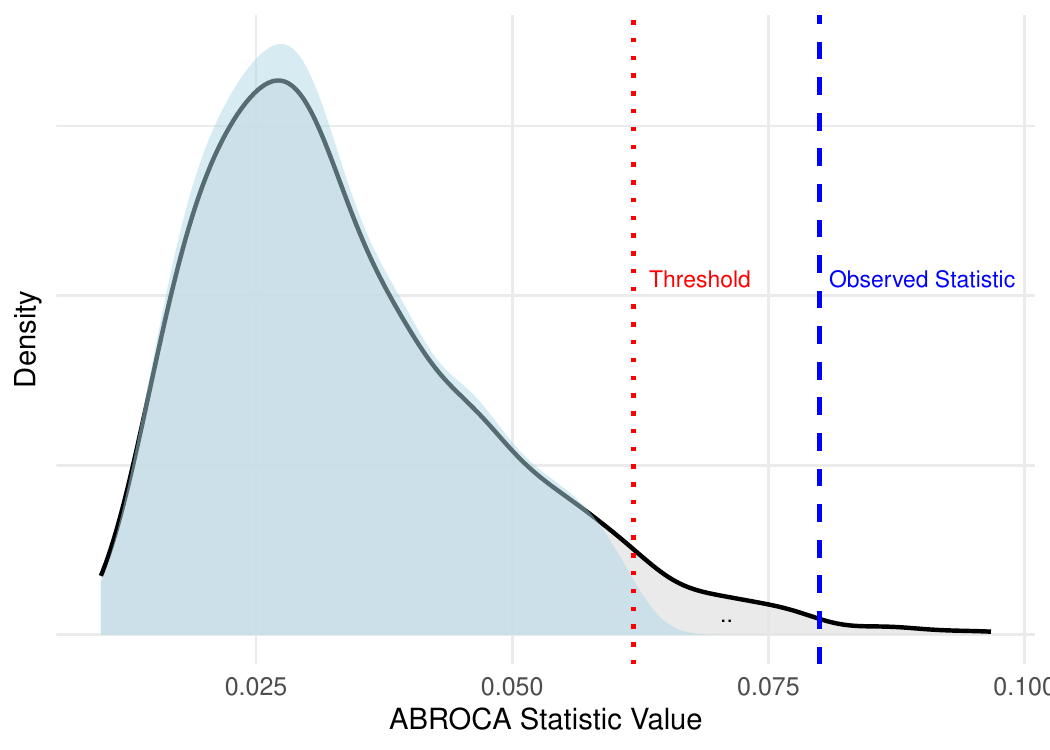}
    \Description{A density plot titled 'Density' showing the distribution of the ABROCA statistic. The x-axis is labeled 'ABROCA Statistic Value' and includes values such as 0.025, 0.050, 0.075, and 0.100. The plot features two labeled components: 'Threshold' and 'Observed Statistic,' indicating key points within the distribution.}
    \caption{Distribution of test statistics under the null hypothesis, demonstrating the randomization-based evaluation of the observed ABROCA metric.}
    \label{fig:testdist}
\end{figure}

For RQ3, we use a stable 0.1 AUC difference as the baseline. We independently vary the imbalance of group membership and outcome class imbalance, considering two scenarios: a highly imbalanced case (90\% vs. 10\%) and a balanced case (50\% vs. 50\%). These variations allow us to systematically evaluate the statistical power of the proposed ABROCA-based randomization test under different conditions.

\section{Results}

We present the results of our simulation analysis and distribution analysis of ABROCA to determine viable statistical tests for this metric and their statistical power.

\subsection{RQ1: ABROCA's Distribution}

\begin{figure*}[ht]
    \centering
    \begin{minipage}[t]{0.495\textwidth}
        \centering
        \includegraphics[width=\textwidth]{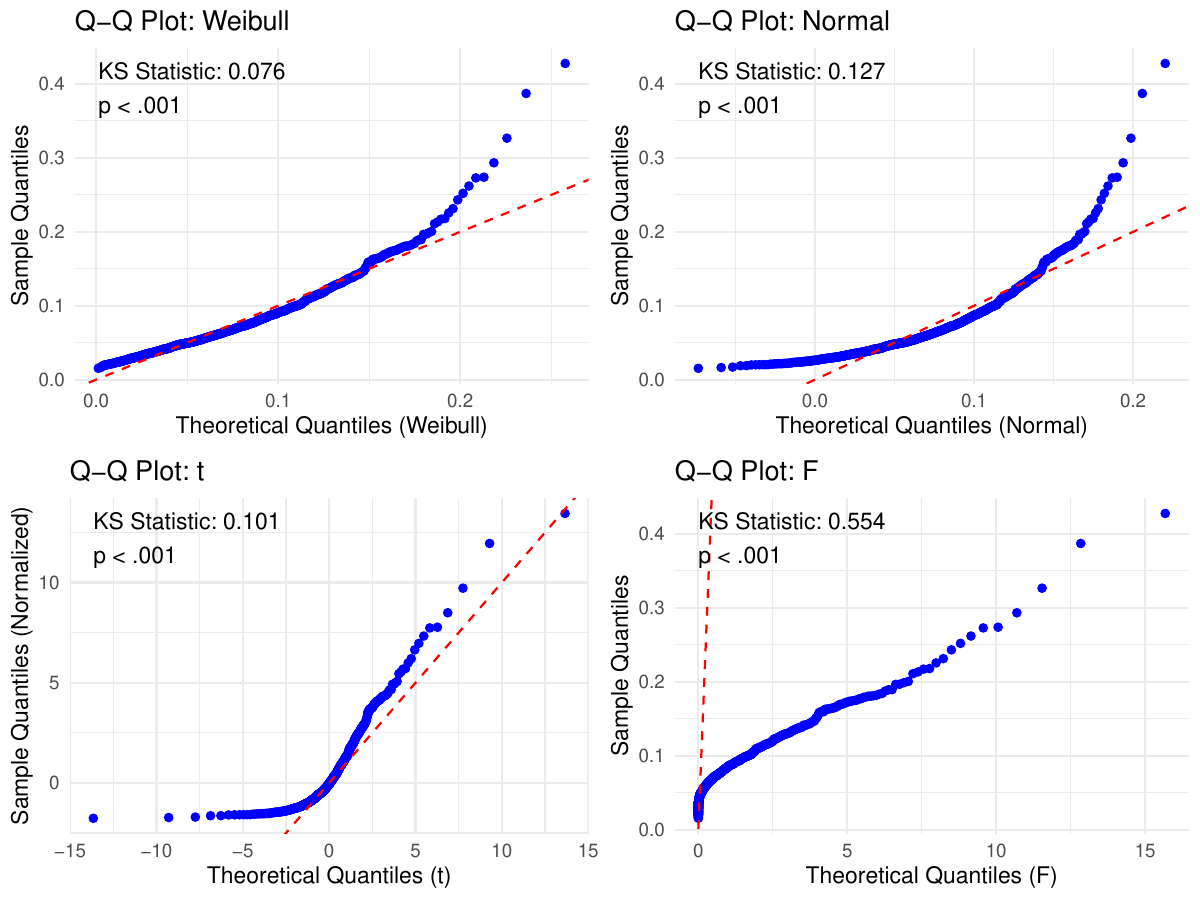}
        \Description{A series of four Q-Q (Quantile-Quantile) plots comparing sample distributions to theoretical distributions for imbalanced data. Each plot is labeled with its type: Weibull, t-distribution, Normal, and F-distribution. Below each plot, the Kolmogorov-Smirnov (KS) statistic and p-value are displayed, indicating the goodness-of-fit (e.g., 'KS Statistic: 0.076, p < .001' for Weibull). The x-axes represent theoretical quantiles, while the y-axes show sample quantiles. The KS statistics suggest significant deviations from the tested distributions, with the largest deviation occurring for the F-distribution (KS = 0.554). The plots are used to assess how well the sample data matches each theoretical model.}
        \caption*{(a) Imbalanced Null Hypothesis}
    \end{minipage}
    \hfill
    \begin{minipage}[t]{0.495\textwidth}
        \centering
        \includegraphics[width=\textwidth]{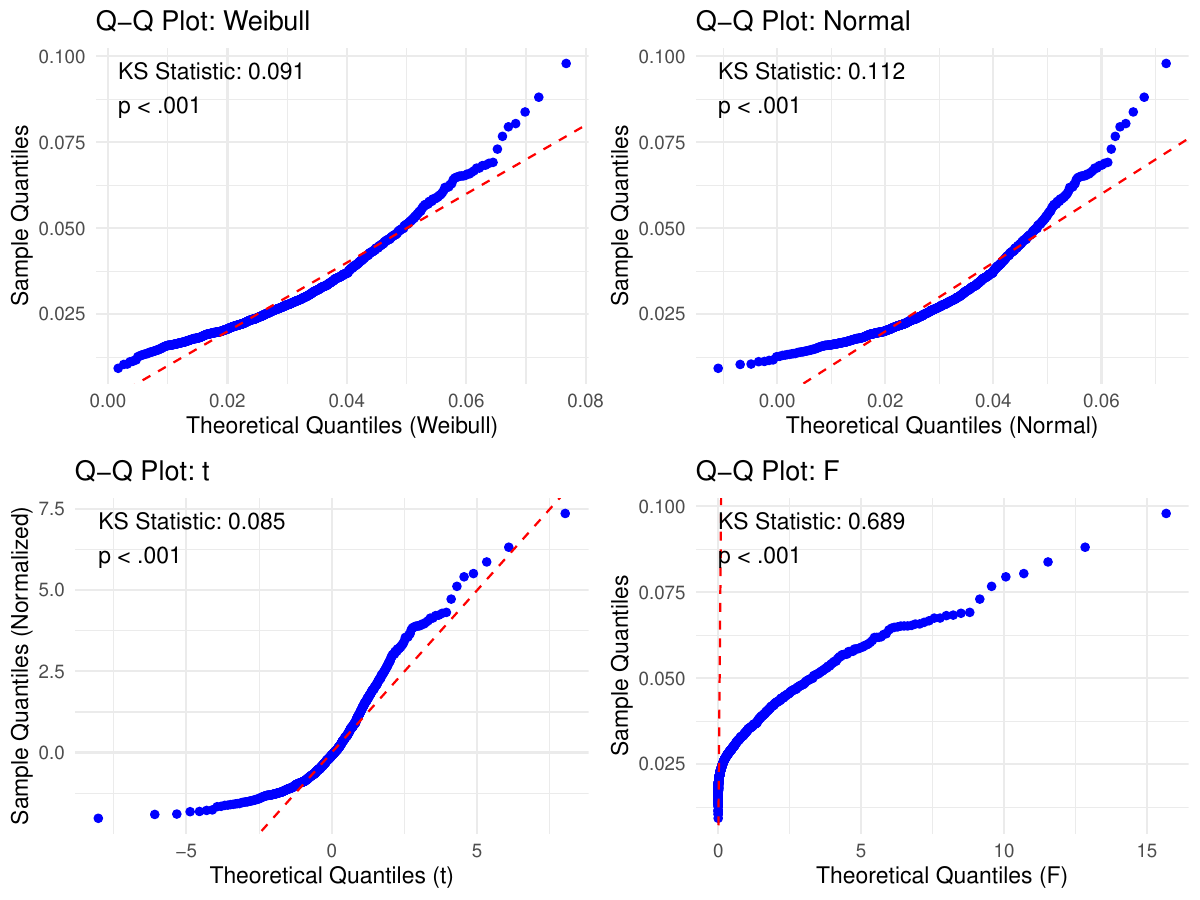}
        \Description{A series of four Q-Q (Quantile-Quantile) plots comparing sample distributions to theoretical distributions for balanced data. Each plot is labeled with its type: Weibull, t-distribution, Normal, and F-distribution. Below each plot, the Kolmogorov-Smirnov (KS) statistic and p-value are displayed, indicating the goodness-of-fit (e.g., 'KS Statistic: 0.091, p < .001' for Weibull). The x-axes represent theoretical quantiles, while the y-axes show sample quantiles. The KS statistics suggest significant deviations from the tested distributions, with the largest deviation occurring for the F-distribution (KS = 0.689). The plots are used to assess how well the sample data matches each theoretical model.}
        \caption*{(b) Balanced Null Hypothesis}
    \end{minipage}
    \caption{Comparison of Q-Q plots for imbalanced and balanced null hypotheses based on a sample size of 5,000. The plots illustrate the deviations of ABROCA sample quantiles from theoretical quantiles for various tested distributions, including Weibull, normal, \(t\)-distribution, and \(F\)-distribution. These comparisons help evaluate which distributions best approximate ABROCA under different class balance conditions. The red line represents the expected relationship for a distribution match.}
    \label{fig:qq-plots}
\end{figure*}

In summary, ABROCA does not conform to any standard known distribution. Among the skew-accommodating and flexible distributions tested, the Weibull distribution fits the data best (but still significantly deviates from the theoretical distribution). As shown in the Q-Q plots (Figure \ref{fig:qq-plots}), the Weibull distribution exhibits deviations at the extreme distribution ends in both balanced and imbalanced scenarios. This breakdown at the tails suggests that even under ideal conditions, the Weibull distribution may not be suitable for accurately modeling ABROCA.

Given these limitations, a randomization test offers a more robust and nonparametric approach to hypothesis testing for ABROCA. Randomization tests avoid assumptions about the underlying distribution and provide flexibility across different scenarios, particularly in the presence of class or group imbalances. Our open-source code facilitates the implementation of randomization tests by simulating ABROCA data points under the null hypothesis, enabling researchers to conduct reliable hypothesis testing without relying on parametric assumptions.

\subsection{RQ2: A Statistical Test for ABROCA}

\begin{figure}[ht]
    \centering
    \includegraphics[width=0.485\textwidth]{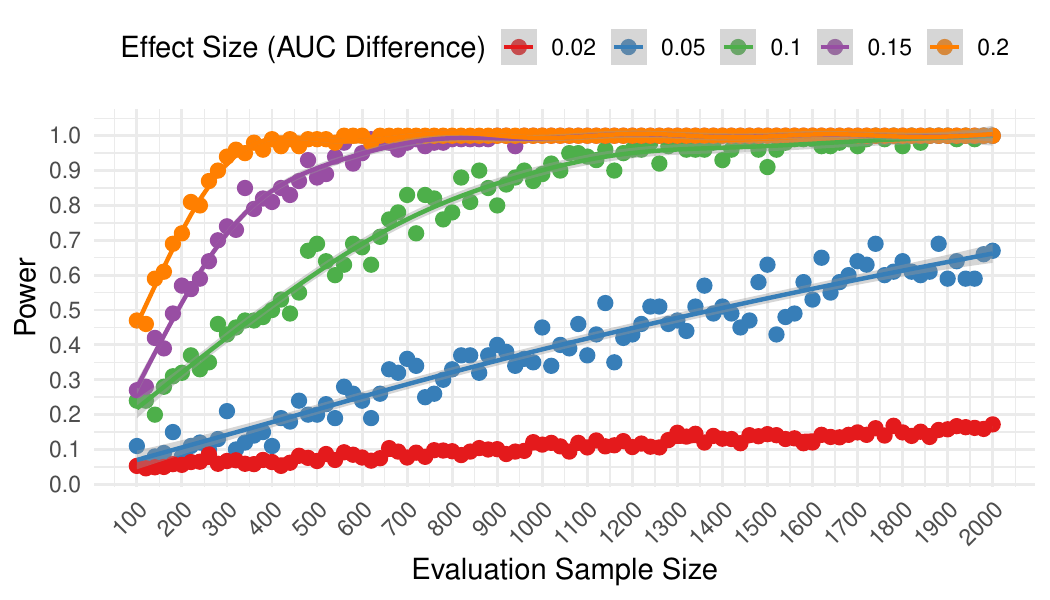}
    \Description{A line graph titled 'Effect Size (AUC Difference)' showing the relationship between evaluation sample size (x-axis) and statistical power (y-axis, ranging from 0.0 to 1.0). The power curve illustrates how the probability of detecting an effect (AUC difference) increases with larger sample sizes. The graph is used to assess the required sample size for achieving desired statistical power (e.g., 0.8 or higher) in a study measuring AUC differences.}
    \caption{Power analysis results showing the relationship between evaluation sample size and statistical power for various effect sizes.}
    \label{fig:power-rq2}
\end{figure}

Figure \ref{fig:power-rq2} illustrates the statistical power achieved for different sample sizes and effect sizes (represented in population-level AUC differences). Notably, for relatively large sample sizes, such as 500 observations per group, or 1,000 holdout test set observations total, achieving the commonly desired statistical power threshold of 0.8 was only possible for an effect size of 0.1 or greater. The power remained well below this threshold for smaller effect sizes (e.g., 0.05 or lower), even with 2,000 total samples.

\subsection{RQ3: Impact of Imbalance on Power}

Our findings indicate that outcome and group class imbalance substantially impact the statistical power of randomization tests to detect differences in ABROCA. When both outcome and group classes were balanced, power was adequate for moderate effect sizes (e.g., an AUC difference of 0.1). However, the power dropped substantially when either group or outcome classes were imbalanced, and especially when both were imbalanced.

\begin{figure}[ht]
    \centering
    \includegraphics[width=0.485\textwidth]{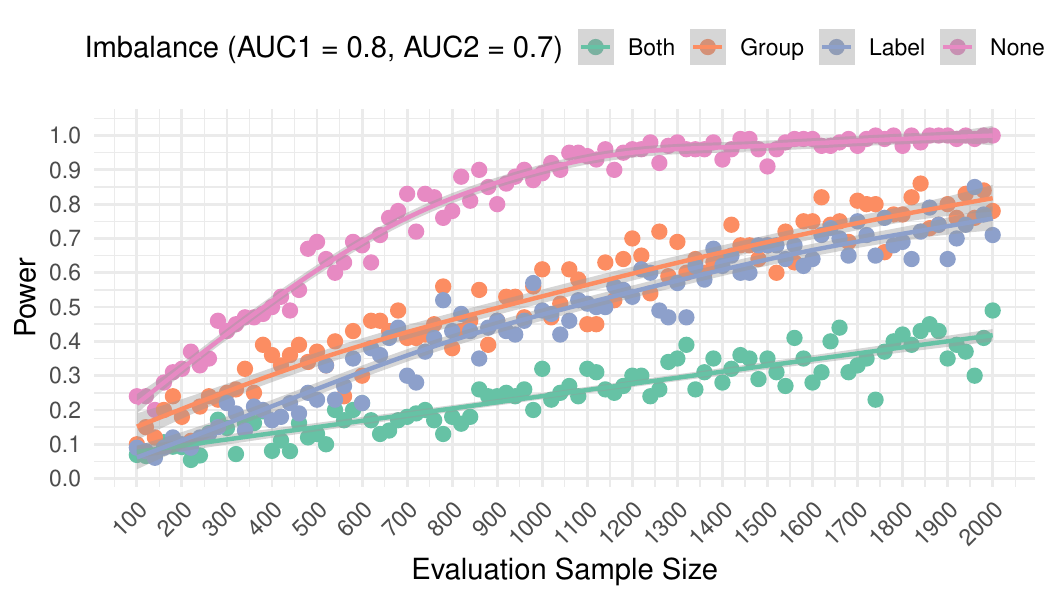}
    \Description{A power analysis graph titled 'Imbalance (AUC1 = 0.8, AUC2 = 0.7)' showing the relationship between evaluation sample size (x-axis, ranging from 100 to 2000) and statistical power (y-axis). The plot compares four data balance or imbalance conditions labeled 'None', 'Both', 'Group', and 'Label', representing different sampling methods. The graph illustrates how power increases with larger sample sizes when detecting an AUC difference between two models (0.8 vs. 0.7). This visualization helps determine the required sample size to achieve adequate statistical power for fairness or performance comparisons.}
    \caption{Power analysis results showing the relationship between evaluation sample size and statistical power for different data imbalances.}
    \label{fig:power-rq3}
\end{figure}

Figure~\ref{fig:power-rq3} illustrates the power curves for varying sample sizes and levels of imbalance. Statistical power remains low even with a test set size of 2,000 when either group or label classes (or both) are imbalanced.

\section{Discussion}

Our findings underscore the challenges in reliably detecting algorithmic bias via the ABROCA metric, especially when group or outcome class distributions are imbalanced. Results suggest that many typical educational data mining (EDM) studies may be at risk of being inconclusive about bias due to low statistical power. The presented findings bear implications for the fairness of deployed models and how future research should be designed to achieve adequate sensitivity to meaningful differences in model performance.

\paragraph{ABROCA's Distribution (RQ1)} 
Simulations confirmed that ABROCA does not conform to any standard distribution (including normal, $t$-, $F$-, and Weibull). Even though Weibull provided a closer fit than the other candidates, it still exhibited systematic deviations, particularly in the distribution tails. Consequently, relying on off-the-shelf parametric tests (e.g., $t$-tests) for ABROCA can be problematic, as standard distributional assumptions are not met. The distributional skewness and heavy tails reinforce that ABROCA requires more specialized inference methods. We propose randomization tests as a nonparametric alternative to testing whether an obtained ABROCA value is significantly above chance, an issue that recent research has identified as a key issue in the widespread use of ABROCA, given its heavy skew \cite{borchers2025abroca}. Researchers and practitioners may use the open-source code presented here to reproduce the randomized testing procedure, including sample size planning through statistical power simulation converging within a few hours on contemporary standard computers.

\paragraph{Statistical Testing and Power (RQ2)} 
We introduced and evaluated a randomization-based approach for testing whether observed ABROCA values correspond to a statistically significant difference in classification performance between groups. Simulation results revealed that small-to-moderate effect sizes (e.g., population AUC differences around 0.05 or less) are challenging to detect with conventional sample sizes common in EDM. For instance, obtaining a power of 0.8 to detect a 0.05 population AUC difference using ABROCA often required well over 2{,}000 labeled instances in the holdout test set. These sample sizes are common in EDM but even smaller when special groups of students, such as minorities or intersectional identities, are evaluated regarding a model's bias \cite{zambrano2024investigating,vsvabensky2024evaluating}.
These findings highlight the ease with which real disparities might go unnoticed (Type II errors) or detected bias is a statistical artifact and not real (Type I errors) and the risk of misevaluating fairness when imprecision is high, especially when employing skewed metrics such as ABROCA \cite{gardner2019evaluating,borchers2025abroca}. More generally, these findings suggest estimating algorithmic bias in minority or intersectional identities with high reliability is very challenging.

\paragraph{Impact of Imbalances (RQ3)} 
Class and group imbalances further diminish the statistical power of ABROCA tests. The obtained results show that even moderate effect sizes become exceedingly difficult to detect when data reflect the real-world disparities in demographic group sizes or outcome prevalence often encountered in educational contexts \cite{zambrano2024investigating}. The lower baseline detectability in imbalanced settings means that researchers may need to sample substantially more data to achieve the same statistical power as in balanced scenarios. Given that imbalances are common in EDM (e.g., minority groups in a student population or small numbers of positive outcomes in dropout prediction), these results underscore the importance of large or oversampled cohorts to avoid underestimating or overestimating bias.

\subsection{Implications}

The presented findings have several implications. First, researchers observing small ABROCA values should consider whether their tests are sufficiently powered before concluding no model bias exists. In underpowered settings, random chance could obscure the presence (or absence) of genuine discrepancies in model performance. Second, lower statistical power may explain conflicting results across studies: if different datasets have varying degrees of group imbalance and sample sizes, some might detect bias while others do not, despite seemingly similar conditions. A lot of algorithmic bias research in EDM has been performed in higher education samples (where holdout test sets, typically 20\% of the sample, rarely exceed 2,000 observations \cite{vsvabensky2024evaluating,zambrano2024investigating}). Therefore, the present results suggest pooling data across multiple study sites may be required to evaluate algorithmic bias reliably. This is especially true when groups (e.g., minority student populations) or outcomes (e.g., dropout prediction) are imbalanced. Finally, practitioners aiming to ensure fair outcomes must pay close attention to sample and effect sizes early in the study design phase. A power analysis can help calibrate whether sufficient data have been collected to make reliable claims about a model’s fairness \cite{wickelmaier2021simulating}.

\subsection{Limitations and Future Work}
\label{sec:disc:lim}

There are several limitations to the present study. First, we focused on holdout test sets rather than the common cross-validation procedure in EDM. Leave-one-out cross-validation is generally asymptotically equivalent to comparing model $AIC$ \cite{shao1993linear}, an alternative criterion to determining whether two models or theories (e.g., that of a biased or unbiased model) describe empirical data better. More research is needed to describe the reliability of algorithmic bias assessments under such research frameworks. Second, like many simulation studies, our results rely on synthetic data-generating processes, which may not capture all the complexities of real-world educational datasets (e.g., missing data, high-dimensional feature spaces, or intricate dependencies) \cite{Kaser2024simulated}. Future research could examine how these intricacies affect bias detection and ABROCA’s distribution using our open-source code. These applications may include setting simulation parameters according to real-world data sets or applying our testing procedure to data directly. Third, we did not propose a specific minimal (practically) relevant effect size for ABROCA. Determining this threshold is domain- and context-dependent, requiring input from practitioners and policy-makers regarding what magnitude of bias is ethically or educationally meaningful. Deliberation of minimally relevant effect sizes during the study design phase is foundational to sample size planning using statistical power simulations \cite{wickelmaier2021simulating}. In our field, these deliberations could consider whether an intervention is fail-soft or high-stakes as well as its cost (where cost could be conceived as economic, ethical, or otherwise), especially in situations where it may be challenging to obtain high-precision prediction models \cite{baker2025difficulty}.

Another important avenue for future work involves examining localized regions within the ROC curve for group-wise disparities. One strength of ABROCA is its capability to reflect differential performance across varying decision thresholds, but our statistical tests currently treat the metric as a single value. Investigating how power changes when focusing on specific areas of the ROC curve (e.g., points relevant to intervention decisions in at-risk student predictions) could yield a more nuanced understanding of model fairness. Finally, expanding these methods to other fairness metrics beyond ABROCA (e.g., MADD, equalized odds) will strengthen our field's toolkit for robust, reproducible bias assessment.

\section{Conclusion}

This study examined the distribution of the ABROCA metric and introduced a nonparametric randomization test to assess its significance for algorithmic bias detection. Our simulations confirmed that ABROCA’s skewness and heavy-tailed behavior make the metric ill-suited to standard parametric tests. We demonstrated that in many realistic scenarios—particularly those involving modest sample sizes and imbalanced group or outcome class distributions—ABROCA-based bias analyses can suffer from low statistical power, potentially leading to inconclusive results about fairness. Consequently, studies might fail to detect genuine model bias or incorrectly infer bias where none exists due to random chance.

By providing open-source code to reproduce our randomization-based significance tests and power analyses, we enable other researchers to evaluate ABROCA in diverse contexts. Our findings underscore the need for a cautious interpretation of ABROCA values: the absence of evidence for bias is not necessarily evidence of fairness if the analysis is underpowered. Researchers should plan for sufficient sample sizes or adopt data pooling and oversampling strategies when working with imbalanced datasets. We hope this work encourages the broader adoption of robust statistical frameworks in algorithmic bias assessment, ultimately supporting more reliable model evaluations and equitable model application outcomes.

\section*{Acknowledgments}

I thank Ryan S. Baker for pointing out the intricate relationship between cross-validation and statistical testing (see Section \ref{sec:disc:lim}) and for earlier discussions that led to the inception of this research.

\bibliographystyle{abbrv}
\bibliography{main}  

\begin{thebibliography}{10}

\bibitem{baker2025difficulty}
R.~Baker, C.~Mills, and J.~Choi.
\newblock The difficulty of achieving high precision with low base rates for high-stakes intervention.
\newblock In {\em Proceedings of the 15th International Learning Analytics and Knowledge Conference}, pages 790--796, 2025.

\bibitem{baker2022algorithmic}
R.~S. Baker and A.~Hawn.
\newblock Algorithmic bias in education.
\newblock {\em International Journal of Artificial Intelligence in Education}, 32:1--41, 2022.

\bibitem{barocas2020hidden}
S.~Barocas, A.~D. Selbst, and M.~Raghavan.
\newblock The hidden assumptions behind counterfactual explanations and principal reasons.
\newblock In {\em Proceedings of the 2020 Conference on Fairness, Accountability, and Transparency}, pages 80--89, 2020.

\bibitem{borchers2025abroca}
C.~Borchers and R.~S. Baker.
\newblock {ABROCA distributions for algorithmic bias assessment: Considerations around interpretation}.
\newblock In {\em Proceedings of the 15th International Learning Analytics and Knowledge Conference}, pages 837--843, 2025.

\bibitem{bowers2019receiver}
A.~J. Bowers and X.~Zhou.
\newblock Receiver operating characteristic (roc) area under the curve (auc): A diagnostic measure for evaluating the accuracy of predictors of education outcomes.
\newblock {\em Journal of Education for Students Placed at Risk (JESPAR)}, 24(1):20--46, 2019.

\bibitem{choi2025bias}
J.~Choi, S.~Karumbaiah, and J.~Matayoshi.
\newblock Bias or insufficient sample size? improving reliable estimation of algorithmic bias for minority groups.
\newblock In {\em Proceedings of the 15th International Learning Analytics and Knowledge Conference}, pages 547--557, 2025.

\bibitem{cock2023protected}
J.~M. Cock, M.~Bilal, R.~Davis, M.~Marras, and T.~Kaser.
\newblock Protected attributes tell us who, behavior tells us how: A comparison of demographic and behavioral oversampling for fair student success modeling.
\newblock In {\em LAK23: 13th International Learning Analytics and Knowledge Conference}, pages 488--498, 2023.

\bibitem{deho2024past}
O.~B. Deho, L.~Liu, J.~Li, J.~Liu, C.~Zhan, and S.~Joksimovic.
\newblock When the past != the future: Assessing the impact of dataset drift on the fairness of learning analytics models.
\newblock {\em IEEE Transactions on Learning Technologies}, 17:1007--1020, 2024.

\bibitem{fitdistRplus}
M.~L. Delignette-Muller and C.~Dutang.
\newblock {fitdistrplus}: An {R} package for fitting distributions.
\newblock {\em Journal of Statistical Software}, 64(4):1--34, 2015.

\bibitem{friedman1996bias}
B.~Friedman and H.~Nissenbaum.
\newblock Bias in computer systems.
\newblock {\em ACM Transactions on Information Systems (TOIS)}, 14(3):330--347, 1996.

\bibitem{gardner2019evaluating}
J.~Gardner, C.~Brooks, and R.~Baker.
\newblock Evaluating the fairness of predictive student models through slicing analysis.
\newblock In {\em Proceedings of the 9th International Conference on Learning Analytics \& Knowledge}, pages 225--234, 2019.

\bibitem{haim2023open}
A.~Haim, R.~Gyurcsan, C.~Baxter, S.~T. Shaw, and N.~T. Heffernan.
\newblock How to open science: Debugging reproducibility within the educational data mining conference.
\newblock In {\em Proceedings of the 16th International Conference on Educational Data Mining}, pages 114--124, 2023.

\bibitem{jeni2013facing}
L.~A. Jeni, J.~F. Cohn, and F.~De~La~Torre.
\newblock Facing imbalanced data: Recommendations for the use of performance metrics.
\newblock In {\em 2013 Humaine Association Conference on Affective Computing and Intelligent Interaction}, pages 245--251. IEEE, 2013.

\bibitem{jiang2021towards}
W.~Jiang and Z.~A. Pardos.
\newblock Towards equity and algorithmic fairness in student grade prediction.
\newblock In {\em Proceedings of the 2021 AAAI/ACM Conference on AI, Ethics, and Society}, pages 608--617, 2021.

\bibitem{karimi2021predicting}
M.~Karimi-Haghighi, C.~Castillo, D.~Hern{\'a}ndez-Leo, and V.~M. Oliver.
\newblock Predicting early dropout: Calibration and algorithmic fairness considerations.
\newblock {\em arXiv preprint arXiv:2103.09068}, 2021.

\bibitem{Kaser2024simulated}
T.~K{\"a}ser and G.~Alexandron.
\newblock {Simulated Learners in Educational Technology: A Systematic Literature Review and a Turing-like Test}.
\newblock {\em International Journal of Artificial Intelligence in Education}, 34(2):545--585, 2024.

\bibitem{kizilcec2022algorithmic}
R.~F. Kizilcec and H.~Lee.
\newblock Algorithmic fairness in education.
\newblock In W.~Holmes and K.~Porayska-Pomsta, editors, {\em The Ethics of Artificial Intelligence in Education}, pages 174--202. Routledge, 2022.

\bibitem{pinho2012gamma}
L.~Pinho, G.~Cordeiro, and J.~Nobre.
\newblock The gamma-exponentiated weibull distribution.
\newblock {\em Journal of Statistical Theory and Applications}, 11(4):379--395, 2012.

\bibitem{sha2021assessing}
L.~Sha, M.~Rakovic, A.~Whitelock-Wainwright, D.~Carroll, V.~M. Yew, D.~Gasevic, and G.~Chen.
\newblock Assessing algorithmic fairness in automatic classifiers of educational forum posts.
\newblock In {\em Artificial Intelligence in Education: 22nd International Conference, AIED 2021, Utrecht, The Netherlands, June 14--18, 2021, Proceedings, Part I}, pages 381--394. Springer, 2021.

\bibitem{shao1993linear}
J.~Shao.
\newblock Linear model selection by cross-validation.
\newblock {\em Journal of the American statistical Association}, 88(422):486--494, 1993.

\bibitem{vsvabensky2024evaluating}
V.~{\v{S}}v{\'a}bensk{\`y}, M.~Verger, M.~M.~T. Rodrigo, C.~J.~G. Monterozo, R.~S. Baker, M.~Z. N.~L. Saavedra, S.~Lall{\'e}, and A.~Shimada.
\newblock Evaluating algorithmic bias in models for predicting academic performance of filipino students.
\newblock In {\em Proceedings of the 17th International Conference on Educational Data Mining}, pages 744--751, 2024.

\bibitem{verger2023your}
M.~Verger, S.~Lall{\'e}, F.~Bouchet, and V.~Luengo.
\newblock {Is Your Model ``MADD'''? A Novel Metric to Evaluate Algorithmic Fairness for Predictive Student Models}.
\newblock In {\em Proceedings of the 16th International Conference on Educational Data Mining}, pages 91--102, 2023.

\bibitem{verma2018fairness}
S.~Verma and J.~Rubin.
\newblock Fairness definitions explained.
\newblock In {\em Proceedings of the International Workshop on Software Fairness}, pages 1--7, 2018.

\bibitem{wickelmaier2021simulating}
F.~Wickelmaier.
\newblock Simulating the power of statistical tests: A collection of {R} examples.
\newblock {\em arXiv preprint arXiv:2110.09836}, 2021.

\bibitem{xu2024contexts}
Z.~Xu, J.~Olson, N.~Pochinki, Z.~Zheng, and R.~Yu.
\newblock Contexts matter but how? course-level correlates of performance and fairness shift in predictive model transfer.
\newblock In {\em Proceedings of the 14th Learning Analytics and Knowledge Conference}, pages 713--724, 2024.

\bibitem{zambrano2024investigating}
A.~F. Zambrano, J.~Zhang, and R.~S. Baker.
\newblock Investigating algorithmic bias on bayesian knowledge tracing and carelessness detectors.
\newblock In {\em Proceedings of the 14th Learning Analytics and Knowledge Conference}, pages 349--359, 2024.

\end{thebibliography}

\end{document}